\def\gmode{} 

 \def\gdriver{dviout}

\newcommand{\dtitle}[1]{\title{ \if \gmode \else
\color{red} Demo mode!\\
comment out \textbackslash def \textbackslash gmode\{demo\} at the header to include figures \color{black}\\
\fi
#1 }}
\ifx\pdfoutput\undefined
\else
 \def\gdriver{}
\fi

\documentclass[\gmode,\gdriver,a4paper,conference]{IEEEtran}
\usepackage{amsmath}

\usepackage{graphicx}
\ifCLASSINFOpdf
\else
\fi
\newcommand{\fnote}[1]{}

\newcommand{\knote}[1]{{\color{red} \bf #1 \color{black}}}

\newcommand{\bnote}[1]{{\color{red} #1 \color{black}}}
\newcommand{\mnote}[1]{}
\newcommand{\onote}[1]{\color{blue} #1 \color{black}}
\newcommand{\snote}[1]{}
\newcommand{\kcut}[1]{}
\newcommand{\ocut}[1]{}
\newcommand{\scut}[1]{}

\ifx\pdfoutput\undefined
\else
 \usepackage{CJKutf8}
 \renewcommand{\fnote}[1]{}
 \renewcommand{\knote}[1]{}
 \renewcommand{\mnote}[1]{}
 \renewcommand{\bnote}[1]{}
 \renewcommand{\onote}[1]{}
\fi

\newcommand{\ie}{{\it i.e.}}
\newcommand{\eg}{{\it e.g.}}

\newcommand{\figref}[1]{{Fig.~\ref{fig:#1}}}

\newcommand{\equref}[1]{Eq.~(\ref{equ:#1})}

\usepackage{amsmath}
\usepackage{amsfonts}
\usepackage{color}

\begin{document}
%
\title{Representing a Partially Observed Non-Rigid 3D Human Using Eigen-Texture and Eigen-Deformation}


\author{
\IEEEauthorblockN{Ryosuke Kimura}
\IEEEauthorblockA{%
Kagoshima University, Japan
}
\and
\IEEEauthorblockN{Akihiko Sayo}
\IEEEauthorblockA{%
Kyushu University, Japan
}
\and
\IEEEauthorblockN{Fabian Lorenzo Dayrit}
\IEEEauthorblockA{%
Nara Institute of Science and Technology, Japan
}
\and
\IEEEauthorblockN{Yuta Nakashima}
\IEEEauthorblockA{%
Osaka University, Japan
}
\and
\IEEEauthorblockN{Hiroshi Kawasaki}
\IEEEauthorblockA{%
Kyushu University, Japan
}
\and
\IEEEauthorblockN{Ambrosio Blanco}
\IEEEauthorblockA{%
Microsoft Research Asia, China
}
\and
\IEEEauthorblockN{Katsushi Ikeuchi}
\IEEEauthorblockA{
Microsoft Corp, USA
}
}


%


\maketitle

\begin{abstract}

    Reconstruction of the shape and motion of humans from RGB-D is a challenging problem, receiving much attention in recent years. 
    Recent approaches for full-body reconstruction use a statistic shape model, which is built upon accurate full-body scans of people in skin-tight clothes, to complete invisible parts due to occlusion.
    Such a statistic model may still be fit to an RGB-D measurement with loose clothes but cannot describe its deformations, such as clothing wrinkles. 
    Observed surfaces may be reconstructed precisely from actual measurements, while we have no cues for unobserved surfaces. 
    For full-body reconstruction with loose clothes, we propose to use lower dimensional embeddings of texture and deformation referred to as eigen-texturing and eigen-deformation, to reproduce views of even unobserved surfaces. 
    Provided a full-body reconstruction from a sequence of partial measurements as 3D meshes, the texture and deformation of each triangle are then embedded using eigen-decomposition. 
    Combined with neural-network-based coefficient regression, our method synthesizes the texture and deformation from arbitrary viewpoints. 
    We evaluate our method using simulated data and visually demonstrate how our method works on real data.
    \end{abstract}



%
\IEEEpeerreviewmaketitle

\section{Introduction}
    \vspace{-0.1cm}

    Reconstructing non-rigid (\ie, moving and deforming) objects, such as people and animals, has wide varieties of applications, such as novel view synthesis \cite{guillemaut2009,collet2015}. 
    Being different from rigid object/scene reconstruction (\eg~\cite{jancosek2011,furukawa2010,besl1992,izadi2011,newcombe2011,whelan2012}), which can cast 3D reconstruction into an alignment problem, captured frames with non-rigid objects must be handled as a sequence instead of as different views of the same scene since their shape may change from one frame to the next. 
    This makes the problem challenging. 

    Some approaches have addressed this problem without using any prior knowledge of the object \cite{amberg2007,li2008,zollhofer2014,newcomb2015} or with using only the assumption of articulated objects (\eg~\cite{li2009}). 
    These approaches have an inherent weakness in synthesizing unobserved shapes and textures, which is critical for some applications because they often require view synthesis from arbitrary viewpoints, even from an unobserved direction. 

    If we know in advance the object we are going to capture, we can make use of some prior knowledge in order to address this issue. 
    A 3D geometry template is one possible source of prior knowledge for full-body reconstruction. 
    Template-based methods basically acquire a shape template of the target before actually capturing it in motion and subsequently fit the template to measurements obtained from cameras or RGB-D sensors \cite{li2009,zollhofer2014}. 
    This approach largely relies on non-rigid 3D registration and may suffer from insufficient constraints over a possible motion of the target objects, which may be trapped in a local minimum far from the global one.

    For capturing humans, in particular, we can use human shape models (\eg~\cite{malleson2013}) as prior knowledge. 
    Particularly, statistic human shape models (\eg~\cite{anguelov2005,chen2013,bogo2015,SMPL2015}) can serve as a strong regularizer on possible variations and deformations of human bodies, such as poses and body shape (\eg~tall, short, slim, and sturdy). 
    These statistical models are trained with a number of full-body measurements. 
    Through reducing the number of parameters, it is more likely to find a local minimum sufficiently close to the optimal, even with a partial measurement.

    Generally, human bodies exhibit non-rigid deformations according to their poses (\eg~bending arms deforms muscles, skin, and clothes), which we call pose-dependent deformations, and statistic models can describe such pose-dependent deformation only partially. 
    Existing datasets, such as \cite{pishchulin2015}, can be used for training a statistic model but contain measurements of people only in skin-tight clothes; therefore, pose-dependent deformations of muscles and skin are encoded in the model, but those of clothes are not. 

    As mentioned above, the key role of such statistic human shape models is to interpolate unobserved surfaces in measurements of human bodies. 
    However, people rarely wear skin-tight clothes in real situations, and the gap between real measurements and ones in the dataset may hinder from plausible interpolation (for example, clothing folds in unobserved volumes may be smoothed out during the fitting process). 
    A statistic model at least cannot fill in unobserved surfaces with clothing folds, which may cause significant visual artifacts in rendered models.

    This paper proposes a method for full-body reconstruction of moving non-rigid 3D objects, primarily of humans, from RGB-D measurements. 
    Our method also uses a statistic model for rough reconstruction. 
    In this sense, ours is similar to the method by Bogo et al.~\cite{bogo2015}, in which they developed a multi-resolution statistic model and a sophisticated technique for fitting that simultaneously optimizes shape, a single set of textures, and a displacement map. 
    They tested their method with people in skin-tight clothes. 


    In contrast, our method is designed to handle people with loose clothes, in which a statistic model does not work very well. 
    The main idea of our method is that estimation of pose-dependent deformations in unobserved surfaces, which are represented by a relatively small number of parameters using PCA analysis. 
    Instead of finding an accurate 3D mesh, we use a rough 3D mesh as the base shape and apply precise deformations to it. 
    By doing this, our method can handle pose-dependent deformations, such as clothing folds. 

    To achieve this, we propose to use the eigen-texture method\cite{nishino2002,Nakashima:BMVC17}, which embeds view- and light-dependent texture representation of each triangle in 3D meshes in low dimensional spaces, provided that the textures and deformations have a certain regularity. 
    Under the assumption that we can measure a human body in various poses from various directions in the course of measuring the person in motion and that deformation is solely dependent on poses, we can synthesize the pose-dependent components in a human body. 
    The main contribution of this paper is summarized as follows:
    \begin{itemize}
    \item 
    We introduce eigen-texturing to textured full-body reconstruction in order to compress texture representation as well as to synthesize textures on unobserved surfaces. 
    \item 
    We propose eigen-deformation, which embeds the displacement between a statistic model and a fully-fitted 3D mesh into a low dimensional space, enabling the displacement estimation of unobserved surfaces with a relatively small number of parameters.
    \item 
    In order to estimate the parameters for eigen-texturing and eigen-deformation for unobserved body parts, we develop a neural network(NN)-based coefficient regression so as to synthesize a texture and deformation for arbitrary poses as well as viewing directions.
    \end{itemize}

\section{Overview}
    
    \vspace{-0.1cm}
    The difficulty in statistical shape model-based full-body reconstruction of a moving human body in loose clothing lies mostly in the reproduction of pose-dependent deformations that are not described by a statistic shape model. 
    Our idea is to represent such deformations by texture and individual displacement of mesh vertices, both of which are embedded into low dimensional spaces (eigen-texture and eigen-deformation). 
    The individual displacements represent the difference between the statistic shape model's mesh and one that fully registered to the measurement and count for relatively large deformations while the texture reproduces the detail. 
    In the same way as the eigen-texture method~\cite{nishino2002,Nakashima:BMVC17}, our system can compress the storage size used for individual textures and displacements by using a small number of eigenvectors and their coefficients. 
    In addition, it can interpolate unobserved surfaces by using the bases for full body.

\begin{figure}[t]
        \includegraphics[width=\linewidth]{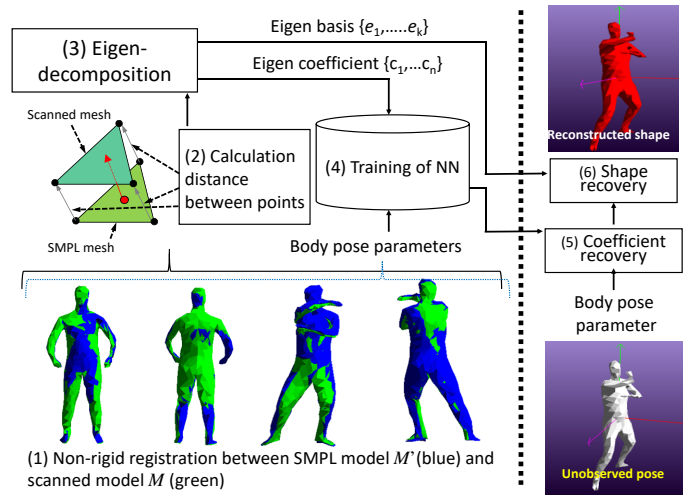}
        \vspace{-0.8cm}
        \caption{Overview of our system. The system first calculates the differences between 
            a scanned shape and a template model, and then, they are represented by small 
            number of eigen basis and coefficients, which are estimated by human pose parameters.}
        \label{fig:overview}
        \vspace{-0.6cm}
        \end{figure}

    \figref{overview} shows an overview of our full-body reconstruction system. 
    At the preprocessing stage, the system registers a statistic shape model to sequences of point clouds obtained from RGB-D measurements. 
    We use SMPL model for non-rigid registration.
    With non-rigid regisitration, we obtain parameters of the statistical model (body shape parameters $v$ and a set of joint angles $\Theta$), as well as mesh $M$ that is fully registered to each point cloud (bottom left of \figref{overview}). 
    As displacements, we compute the difference between $M'$ and $M$, which can be done solely from $\Theta$.
    Then, the displacements are embedded into low dimensional subspace (eigen-deformation) by using a similar way to the eigen-texture method (top left of \figref{overview}).
    At the rendering stage, $M'$ is firstly recovered from $\Theta$, and then the textures and the displacements are reconstructed from the coefficients, which are estimated by $\Theta$ using a NN-based regression.
    Adding all the displacements to $M'$, clothing mesh $M$ is reconstructed.

\vspace{-0.2cm}
\section{Non-rigid registration with SMPL}
    
    \vspace{-0.4cm}
    We used the SMPL model~\cite{SMPL2015} for non-rigid regisitration to simulation data and scanned data.
    First, we make a few correspondence points used as anchor points between SMPL model ($x'_i$) and the simulation or scanned data ($x_i$) manually.
    According to the anchor points, we fit the SMPL model, which is used as the naked model in the regression stage as in \figref{SMPL}(the second model from right), by estimating the pose of it.
    Then, we obtain additional correspondence points using boundaries, where we calculate closest points. 
    Using those correspondences, the SMPL model is inflated to fit the target model. 
    If this step is abandoned, the SMPL model is frequently collapsed due to large shape differences at initial registration (\eg left side of arm corresponds to the right side of arm).
    Finally, we super-sample the vertices and made correspondences between them ($y'_{j}$ and $y_{j}$) with the nearest neighbor method as in \figref{SMPL}(the right-most model).
    After these steps, we conduct non-rigid registration by minimizing following energy function:
    \begin{eqnarray}
    E &=& \alpha \sum_{j \in C_{anc}} \| x'_{j} - x_{j} \| ^2 + \beta \sum_{j \in C_{cor}} \| y'_{j} - y_{j} \| ^2 + P(\Theta) \nonumber \\ 
    &+& \sum_{j \in C_{joint}} R_j(\Theta) + S(v) + \sum_{j \in C_{ver}} \| s_{j-1} - s_{j} \| ^2 , \label{equ:SMPLfitting}
    \end{eqnarray}
    where $C_{anc}$ is the set of anchor point indices, $C_{cor}$ is the set of corresponding point indices, which are made in final step, $C_{joint}$ is the set of all joint indices, $C_{ver}$ is the set of vertex indices, $s_{k}$ is the k-th vertex of the SMPL model, $P$ is penalty for inplausible joint angles of elbow and knee, $R$ is penalty for large changes in joint angles, $S$ is penalty for large $v$ values, and $\alpha$ and $\beta$ are weight values.

\vspace{-0.1cm}
\section{Eigen-texture}
    \vspace{-0.1cm}

    One of the widely accepted ways to reproduce deformations, especially small ones, is to use texture mapping. 
    With the assumption of skin-tight clothes as in Bogo et al.~\cite{bogo2015}, the texture can be a static image; however, we relax this assumption using dynamically changing textures. 
    As a representation of texture, we use eigen-decomposition method for reducing the storage size required for texture images. 
    In addition, our body scans obtained from two RGB-D sensors usually have some unobserved areas. 
    The system reconstructs plausible full-body mesh thanks to the statistic shape model; however, the textures of triangles in such areas are not recoverable due to unavailability of prior knowledge on the texture. 
    Eigen-texture finds the manifold on which textures on the same triangle lie and thus can synthesize the texture for unobserved regions. 

    To extract the texture of a triangle, we first assess the visibility of each triangle in mesh $M$. 
    The system renders $M$ with the standard OpenGL pipeline to get the depth map in the camera coordinate system. 
    We also render a triangle ID map by color-coding the ID of each triangle. 
    For each pixel with a certain triangle ID, the pixel is back-projected onto the corresponding triangle in $M$ and then is projected to the depth map. 
    The pixel should be visible if the difference between the corresponding depth value in the depth map and the third coordinate of the point on the triangle is small with a certain threshold. 
    We judge that a triangle is visible if all pixels in the triangle are visible since partly occluded triangle may significantly spoil the reconstruction. 
    For each visible triangle $n$, 3D positions $v_{n0}$, $v_{n1}$, and $v_{n2}$ of its vertices are projected onto the RGB image to extract the texture for the triangle. 
    Hereinafter the subscript $n$ may be omitted as long as it is not ambiguous.

    Then, we apply eigen-decomposition~\cite{nishino2002} to each triangle, which is briefly introduced here to make the paper self-contained. 
    Let $P$ be a matrix whose $f$-th column $p_f$ is the vectorized texture of a certain triangle that is visible, where $P$ is in $\mathbb{R}^{3N \times F }$, $N$ is the number of pixels in the triangle, and $F$ is the number of frames in which the triangles are visible. 
    $N$ can be arbitrarily set to a sufficiently large number because we can arbitrarily warp the texture. 
    All column vectors in $P$ can be centralized, \ie, $\bar{p}_f = p_f - \bar{p}$, to form $\bar{P}$ using the averaged texture $\bar{p}$ over the frames with being the triangle visible. 
    We can factorize $\bar{P}\bar{P}^\top$ as a following equation: 
    \begin{equation}
    \bar{P}\bar{P}^\top e_i = \lambda_i e_i,
    \end{equation}
    where $\lambda_i$ is the $i$-th eigenvalue and $e_i$ the $i$-th eigenvector. 
    %
    %
    %
    %

    We can embed a texture $p$ into the subspace spanned by a subset of eigenvectors. 
    The low-dimensional representation (\ie, the coefficient for each eigenvector) of texture $p$ can be computed by
    \begin{equation}
    c = (p - \bar{p}) E,
    \end{equation}
    where $E \in \mathbb{R}^{3N\times L}$ is a matrix whose columns are a set of largest eigenvectors ($L$ is the number of eigenvectors for the subspace). 
    We can also reconstruct texture $p$ from $c$ with
    \begin{equation}
    p = E c + \bar{p}.
    \end{equation}
    This means that unobserved textures can be synthesized if we can regress the low-dimensional coefficient vector.

    \begin{figure}[t]
        \begin{tabular}{cc}

        \begin{minipage}{0.40\hsize}

                \tabcolsep = 0.2mm
                \begin{tabular}{ccc}
                \includegraphics[height=0.11\textheight]{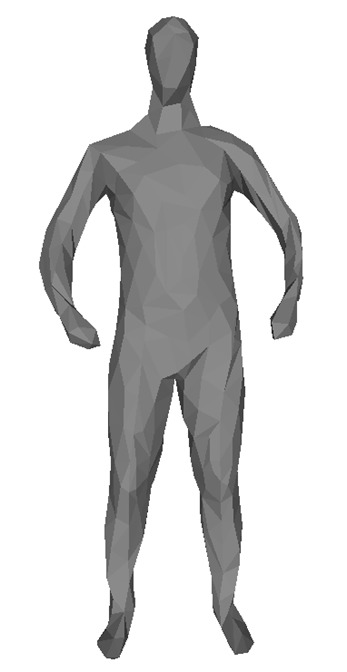} &
                \hspace{-0.3cm}
                \includegraphics[height=0.11\textheight]{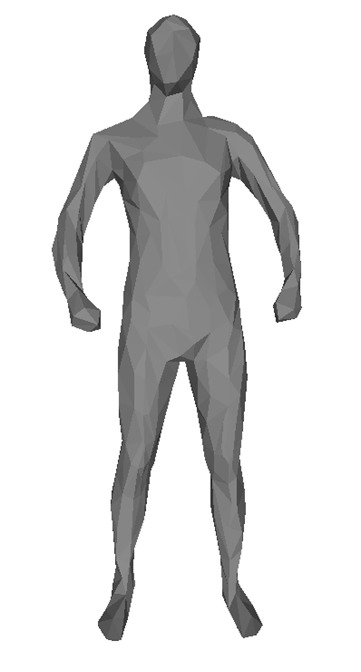} &
                \hspace{-0.3cm}
                \includegraphics[height=0.11\textheight]{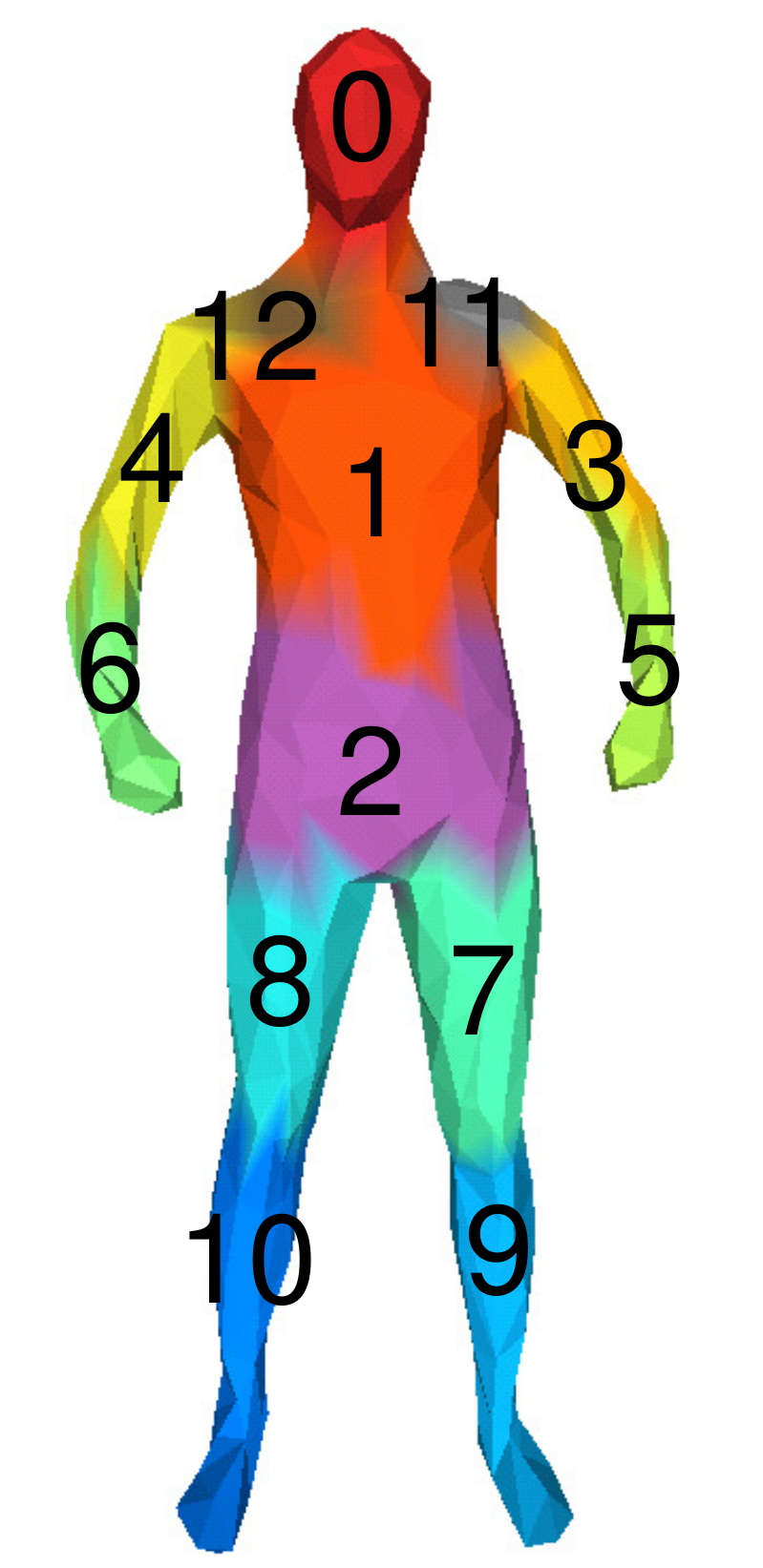} \\
                \end{tabular}
            \vspace{-0.3cm}
            \caption{(Left) clothed model, (middle) TenBo model and (right) body parts 
                segmentation with parts ID.}
            \label{fig:two_models}
            \end{minipage}

        &
        \hspace{-0.2cm}
        \begin{minipage}{0.5\hsize}
                \tabcolsep = 0.2mm
                \begin{tabular}{cccc}
                \includegraphics[height=0.11\textheight]{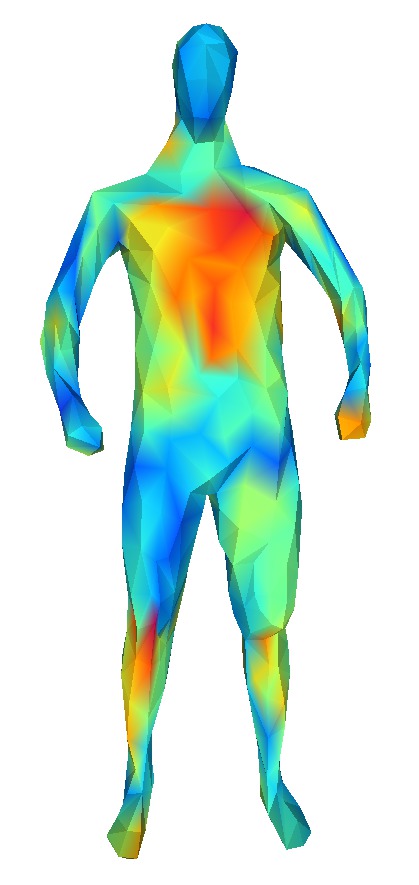} &
                \includegraphics[height=0.11\textheight]{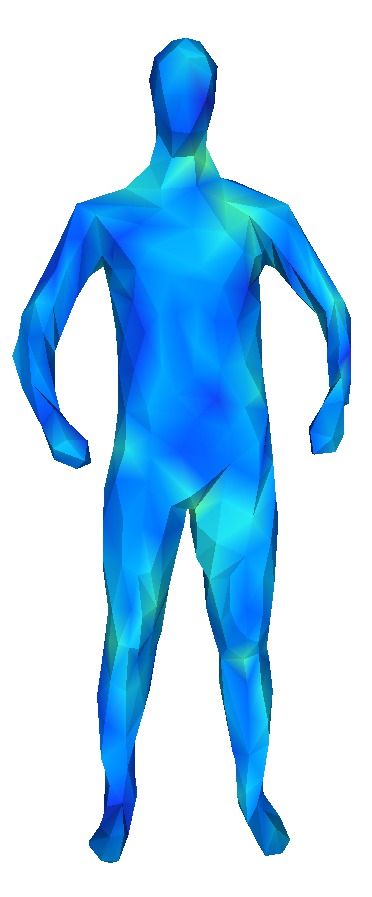} &
                \includegraphics[height=0.11\textheight]{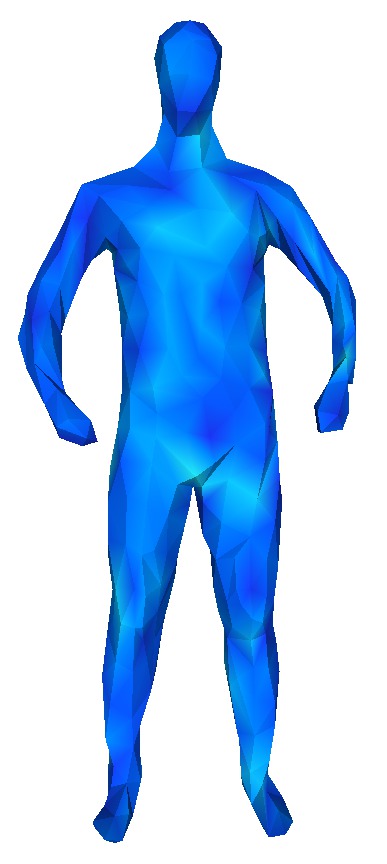} &
                \includegraphics[height=0.1\textheight]{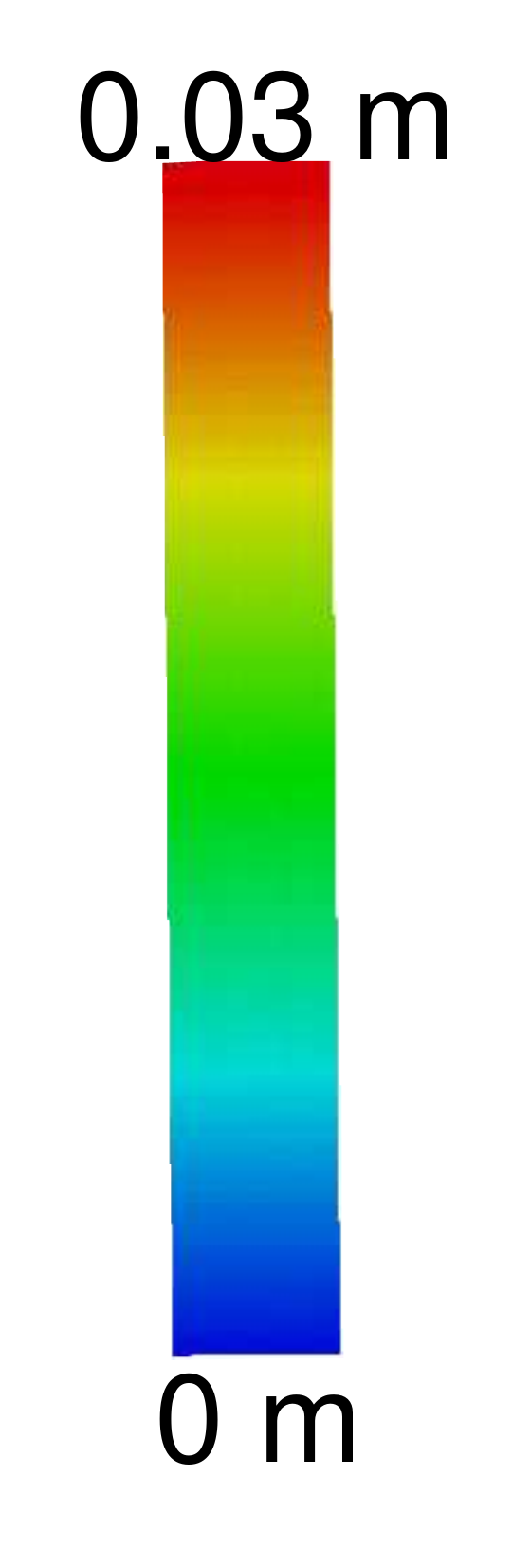}\\
            3 bases&
            5 bases&
            10 bases\\
                \end{tabular}
            \vspace{-0.1cm}
            \caption{RMSE of eigen deformation using a different number of components; 10 components are sufficient for error free reconstruction.}
            \label{fig:PCA_body_real}
            \end{minipage}
        \\

        \end{tabular}   
        \end{figure}

    \begin{figure}[t]
        \centering
        \vspace{-0.6cm}
        \includegraphics[width=0.9\linewidth]{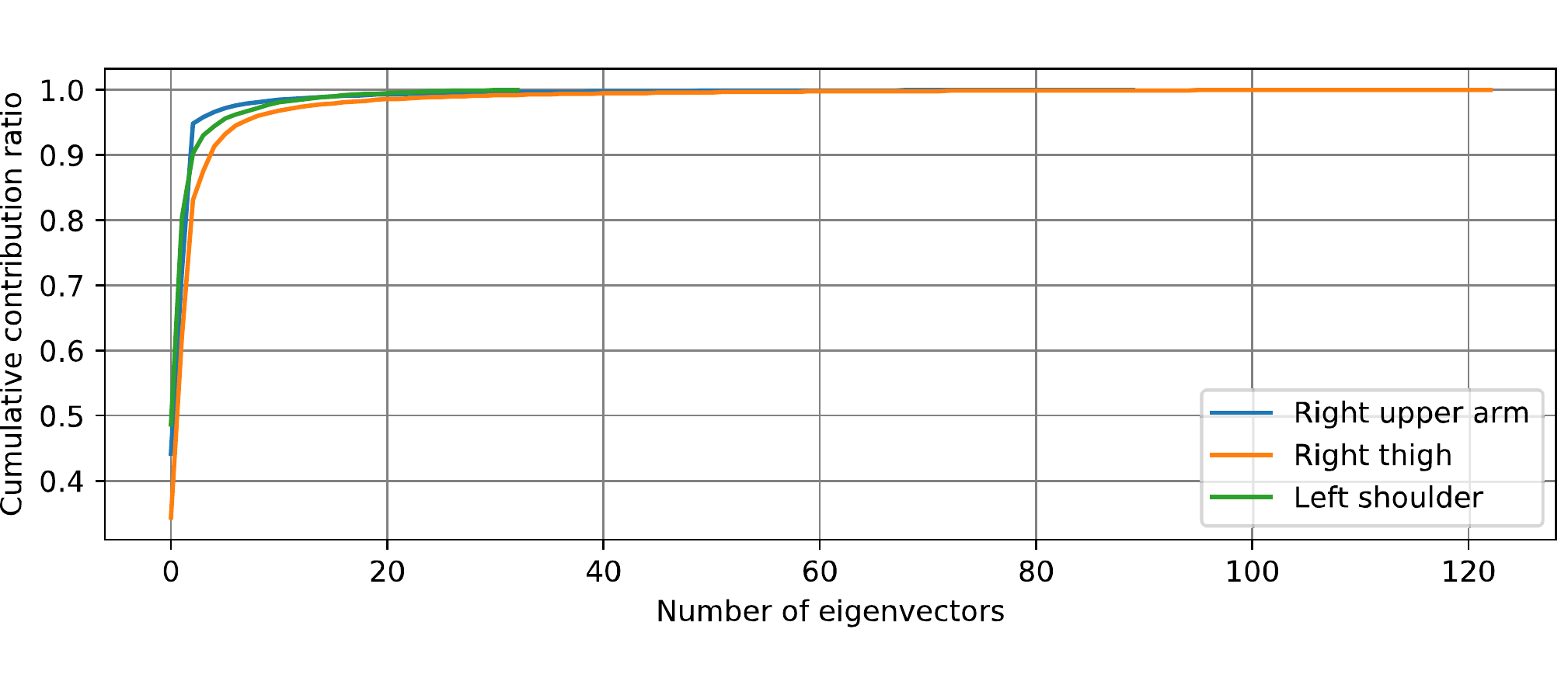}
        \vspace{-0.5cm}
        \caption{Cumulative contribution ratio for three example body parts of real 
            human showing only small number of eigen basis contributes to the original deformation.}
        \vspace{-0.5cm}
        \label{fig:PCA_graph_real}
        \end{figure}

\vspace{-0.1cm}
\section{Eigen-deformation}
    \vspace{-0.1cm}

    Our eigen-texture method is a powerful tool to visually represent deformations caused by loose clothes. 
    However, such texturing-based compensation may not be enough for big deformations. 
    In order to synthesize deformations in unobserved surfaces as well as to compress the storage for storing individual vertex positions, we propose eigen-deformation, inspired by the eigen-texture method.

    The basic idea is almost the same as eigen-texture, but eigen-deformation deals with vertex positions. 
    There is an inherent difference between textures and deformations: Given that the mesh is well registered to the point cloud, the variations in textures are not significant since such variations are caused solely by local deformations, such as wrinkles. 
    On the other hand, those in vertices come from body poses. 
    That is, changes in, \eg, the shoulder joint angle, ends up with large changes in the vertex positions of forearms. 
    Thus, direct application of eigen-decomposition to vertex positions may not work well. 

    To improve the representativity, we compute displacement vectors of each part between the statistic model mesh $M'$ and the fully-registered mesh $M$. 
    We represent the displacement vector in a certain coordinate system associated with each body part; therefore, only the difference between $M'$ and $M$ is counted in the displacement vector.
    Body parts are divided at each joint, and their indices are shown in \figref{two_models}(right).

    The displacement vector $q_{k}$ of the $k$-th vertex of the $l$-th body part in mesh $M$ are computed by $q_{lk} = H_l(v_{lk} - v'_{lk})$ for the corresponding vertex position in $M'$, where $v_{lk}$ and $v'_{lk}$ are the $k$-th vertices in the $l$-th body parts in $M$ and $M'$, and $H_l$ is a rigid transformation matrix between the entire body coordinate system and each body part's coordinate system. 
    We concatenate these displacement vectors to form a column vector $q_l^\top= (q_{l1}^\top \; q_{l2}^\top \; \dots)$, and then aggregate these displacement vectors over all frames in which the triangle is visible. 
    These vectors are centralized as in eigen-texture and again concatenated to form matrix $\bar{Q_l}$. 
    We apply eigen-decomposition to $\bar{Q_l}\bar{Q_l}^\top$ to obtain eigenvectors. 
    We can also embed/reconstruct the displacement vectors into/from the subspace spanned by the eigenvectors.
    Examples of cumulative contribution ratios for three body parts from measurements of a real human body are shown in \figref{PCA_graph_real}. 
    We can see that the ratio drastically increases with a small number of eigenvectors.

    When reconstructing $M$, we firstly recover $M'$ from $\Theta$ and $v$ using \equref{SMPLfitting}, and then recover $\tilde{q_{k}}$ using a small number of eigenvectors. 
    Finally $H_l^{-1}\tilde{q_{k}}$ is calculated for each vertex of each body part and added to $M'$.
    Examples of recovered shapes with a small number of eigenvectors of a real human body are shown in \figref{PCA_body_real}, where the error in recovered shape is visualized by pseudo color, demonstrating a larger number of eigenvectors decrease the error.

\vspace{-0.1cm}
\section{NN-based coefficient regression}
    \vspace{-0.15cm}

    In order for full-body reconstruction, we interpolate unobserved textures and deformations in unobserved surfaces. 
    We do this via coefficient regression in the eigen-texture and eigen-deformation's spaces. 
    Provided that the illumination and RGB-D sensors are fixed in our case, the variations of the textures and the displacements of the same triangle are solely explained by the person's pose. 
    More specifically, joint angles represented by rotation matrices mostly determine them. 
    This implies that coefficients for eigenvectors (or coordinates in the low-dimensional spaces) can be regressed from the rotation matrices. 
    Therefore, we train NN-based regressors that map a joint angle (a rotation matrix) to the coefficients.

\begin{figure}[t]
        \centering
        \includegraphics[width=0.9\linewidth]{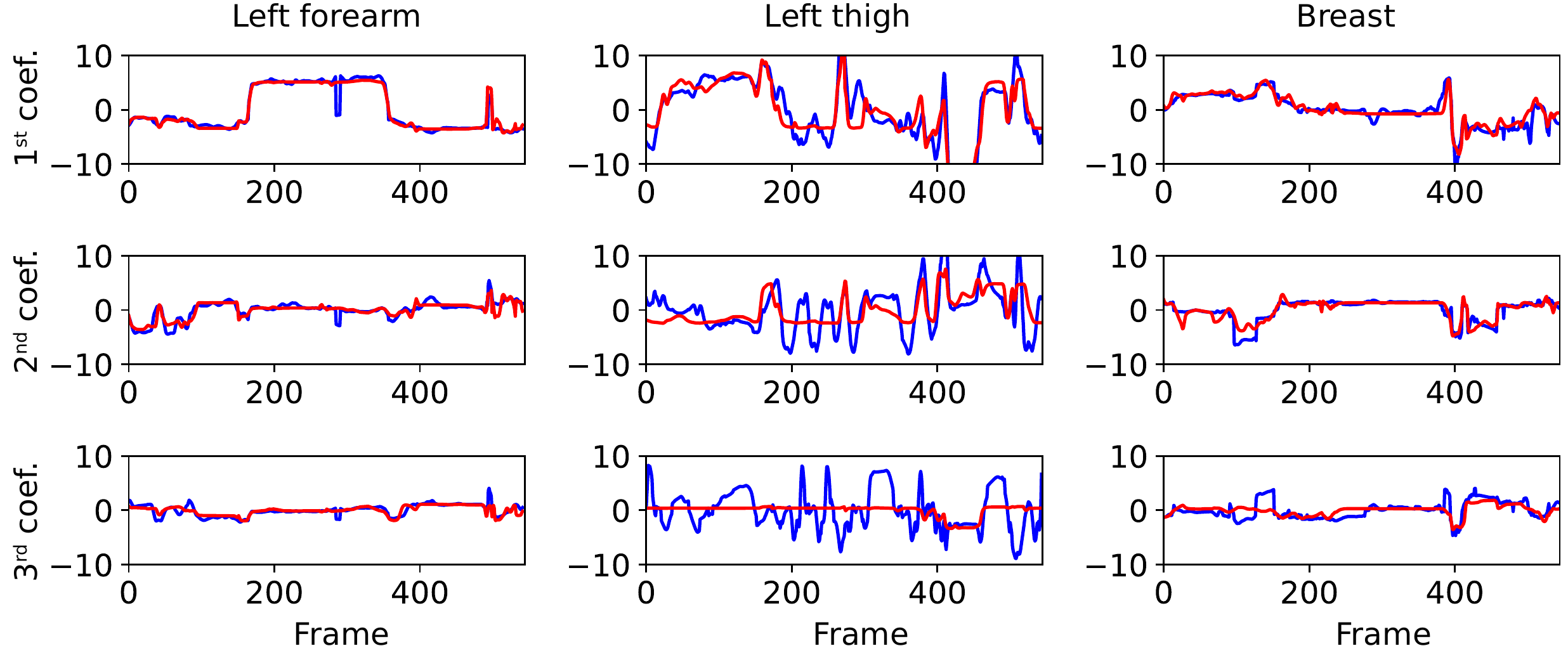}
        \vspace{-0.3cm}
        \caption{Coefficient of eigen value of original data and estimated data.}
        \label{fig:Regress_graph_real}

        \vspace{0.1cm}

            \begin{tabular}{cccc}
            \includegraphics[height=0.13\textheight]{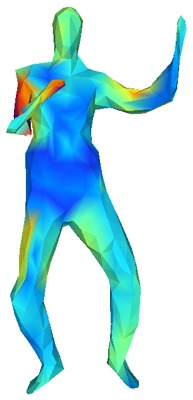} &
            \includegraphics[height=0.13\textheight]{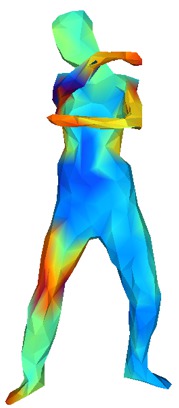} &
            \includegraphics[height=0.13\textheight]{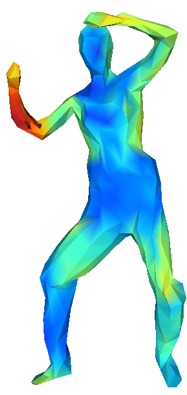} &
            \includegraphics[height=0.10\textheight]{karate-eigen-colorbar-f-m.pdf}\\
            \end{tabular}
        \vspace{-0.1cm}
        \caption{RMSE of eigen deformation using 10 regressed coefficients. Although some 
            parts have large errors, they are come from non-rigid 
            registration failure, and thus, little artifacts observed in final 
            rendering results with texture shown in \figref{Texture_body_real}.}
        \label{fig:Regress_body_real}
        \vspace{-0.3cm}
        \end{figure}

    Let $r \in \mathbb{R}^{9}$ be the vectorization of the rotation matrix that represents a certain body part's joint angle. 
    Since the relationship between rotation matrices and coefficients are unknown, we use a NN with two layers to represent the nonlinearity. 
    Our regressor gives coefficients $\tilde{c} \in \mathbb{R}^L$ by
    \begin{equation}
    \tilde{c}(r) = W_2 \tanh(W_1 r + b_1) + b_2,
    \end{equation}
    where $W_1$ is in $\mathbb{R}^{J \times 9}$ and $W_2$ in $\mathbb{R}^{L \times J}$. The regressor is trained with the gradient descent algorithm. 
    For regularization, we employ weight decay.  
    Examples of estimated coefficient from pose are shown in \figref{Regress_graph_real} and reconstructed shapes are shown in \figref{Regress_body_real}, implying authentication of the algorithm.

\vspace{-0.1cm}
\section{Experiments}
    \vspace{-0.1cm}

    \begin{figure}[t]
        \begin{center}
            \begin{tabular}{cccc}
            \includegraphics[height=0.13\textheight]{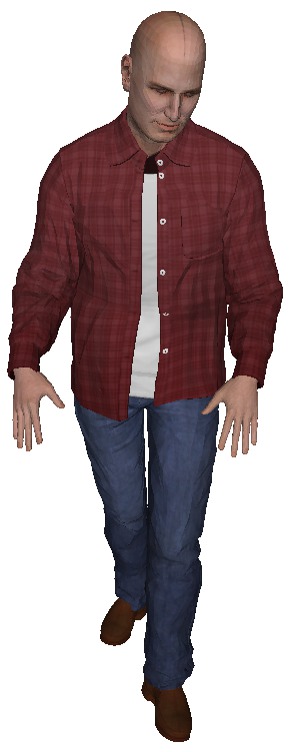} &
            \includegraphics[height=0.13\textheight]{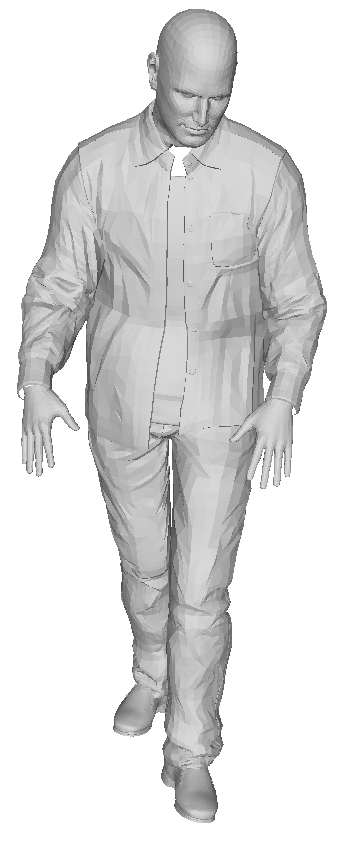} &
            \includegraphics[height=0.13\textheight]{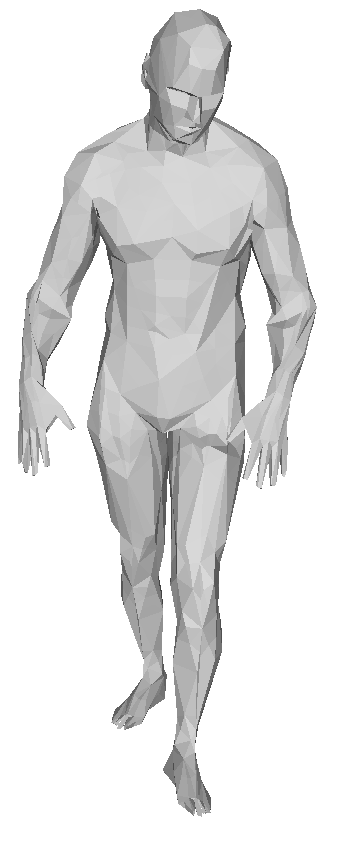} &
            \includegraphics[height=0.13\textheight]{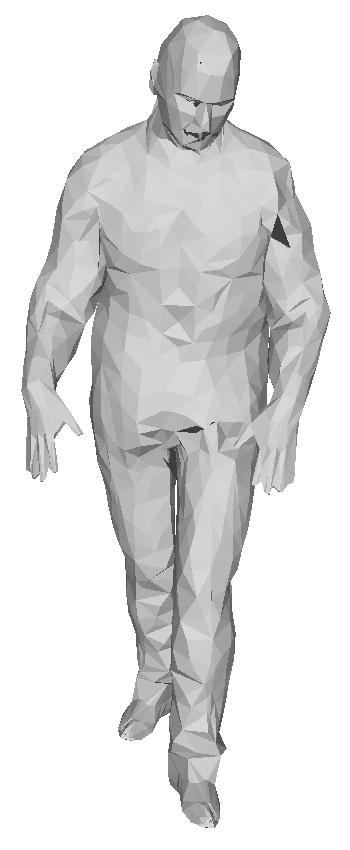}\\
            Textured&
            Non-textured&
            Naked&
            Clothed\\
            model&
            model&
            model&
            model\\
            (sim data)&
            (sim data)&
            (SMPL)&
            (SMPL)\\
            \end{tabular}
        \caption{Synthetic data for evaluation.}
        \label{fig:SMPL} 
        \end{center}    
        \vspace{-0.7cm}
        \end{figure}

    \begin{figure}[t]
        \begin{center}
            \tabcolsep = 0.2mm
            \begin{tabular}{cccc}
            \includegraphics[height=0.13\textheight]{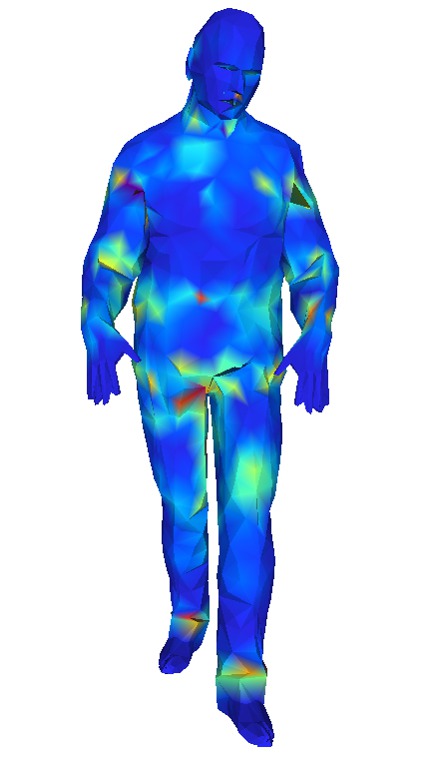} &
            \includegraphics[height=0.13\textheight]{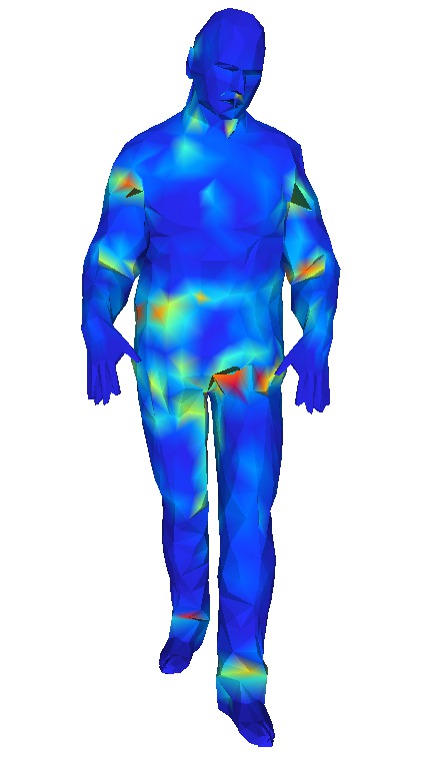} &
            \includegraphics[height=0.13\textheight]{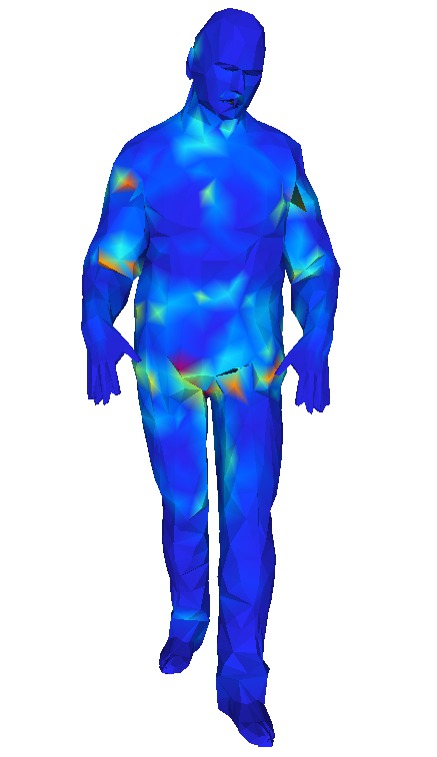} &
            \hspace{-0.5cm}
            \includegraphics[height=0.13\textheight]{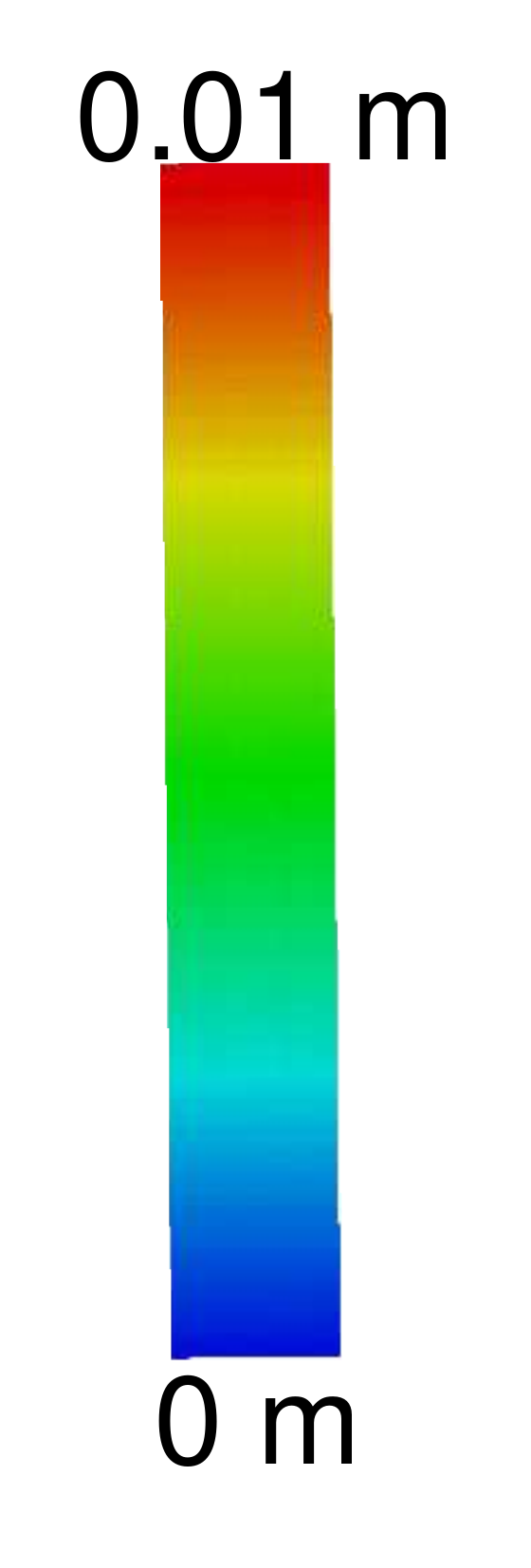}\\
            3 bases&
            5 bases&
            10 bases\\
            \end{tabular}
        \caption{RMSE of eigen deformation using a different number of components; 10 
            components are sufficient for error free reconstruction.}
        \label{fig:PCA_body_sim}
        \end{center}
        \vspace{-0.7cm}
        \end{figure}

    \begin{figure}[t]
        \begin{center}
            \tabcolsep = 0.2mm
            \begin{tabular}{cccc}
            \includegraphics[height=0.13\textheight]{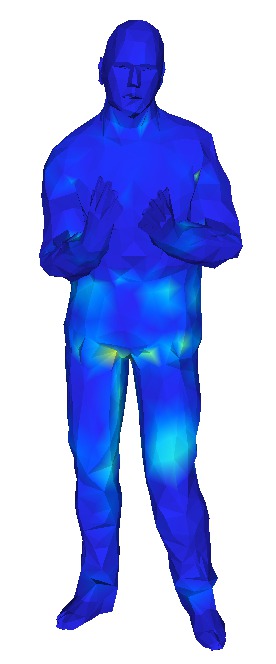} &
            \includegraphics[height=0.13\textheight]{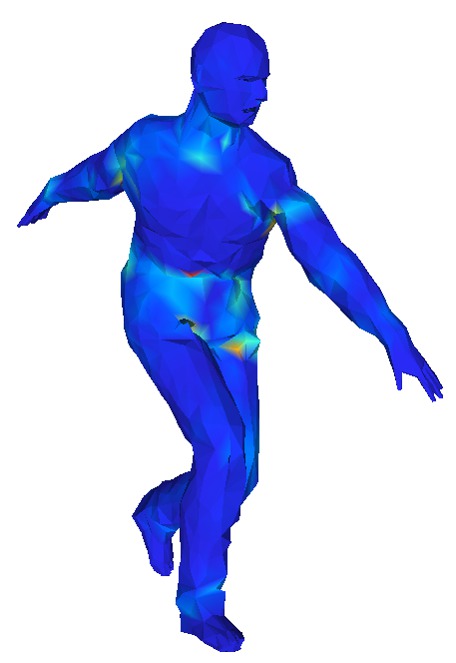} &
            \includegraphics[height=0.13\textheight]{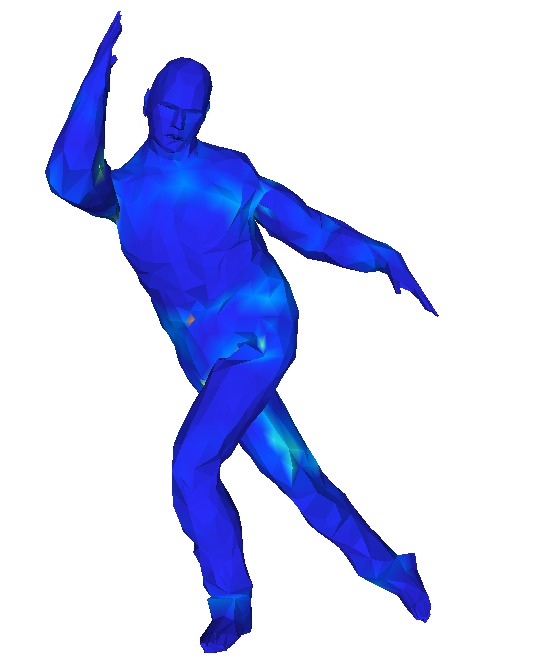} &
            \hspace{-0.5cm}
            \includegraphics[height=0.15\textheight]{sim-eigen-colorbar-m.pdf}\\
            Frame \#300&
            \#400&
            \#440\\
            \end{tabular}
        \caption{RMSE of eigen deformation using 10 regressed coefficients. 
        Unlike the real data (\figref{Regress_body_real}), all errors are 
            small because simulation data has little failure on non-rigid registration.}
        \label{fig:Regress_body_sim}
        \end{center}
        \vspace{-0.8cm}
        \end{figure}

    \begin{figure}[t]
        \centering
        \vspace{-0.4cm}
        \includegraphics[width=0.9\linewidth]{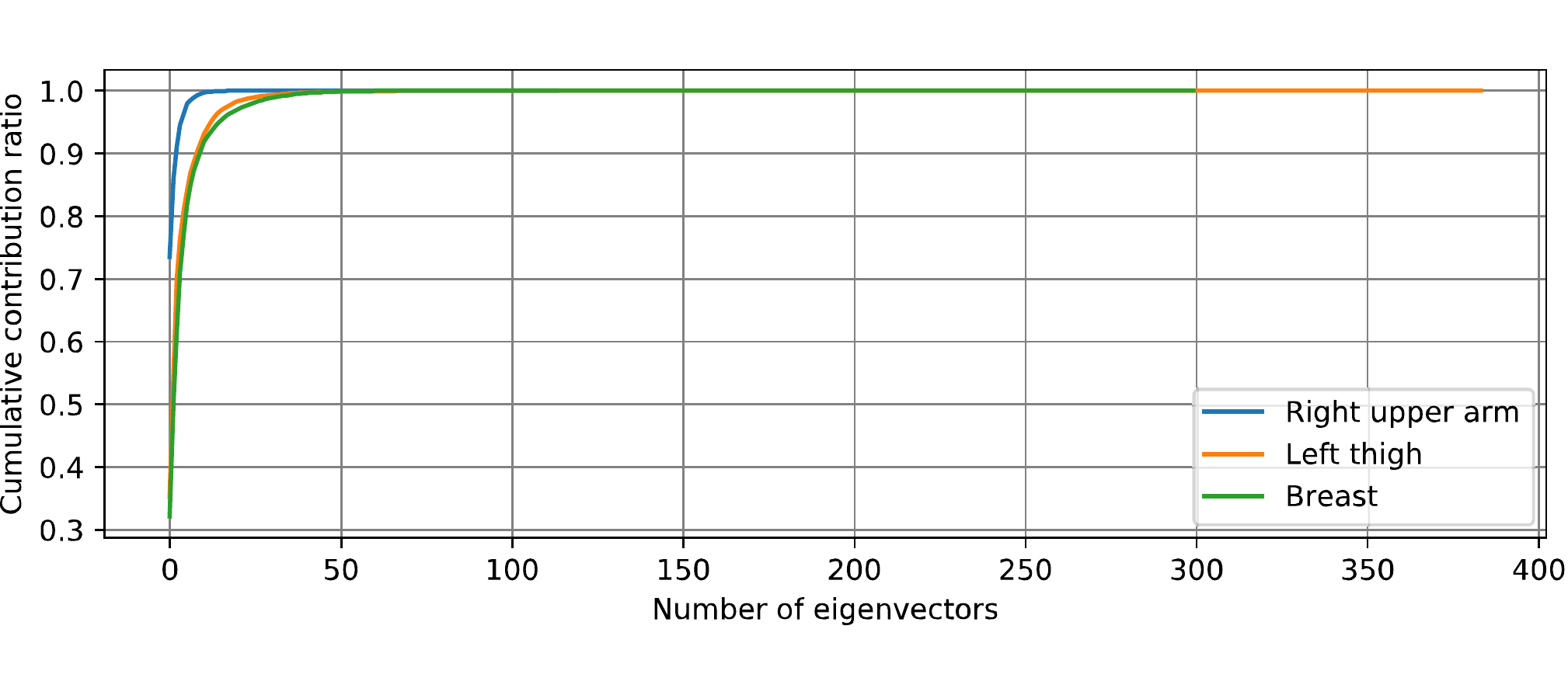}
        \vspace{-0.3cm}
        \caption{Cumulative contribution ratios of three example body parts showing only small 
            number of eigen basis contributes to the original deformation.}
        \label{fig:PCA_graph_sim}
        \vspace{-0.3cm}
        \end{figure}

    \begin{figure*}[ht]
        \begin{tabular}{ccc}
        \vspace{-0.3cm}

        \begin{minipage}{0.25\hsize}
            \includegraphics[width=1.0\textwidth]{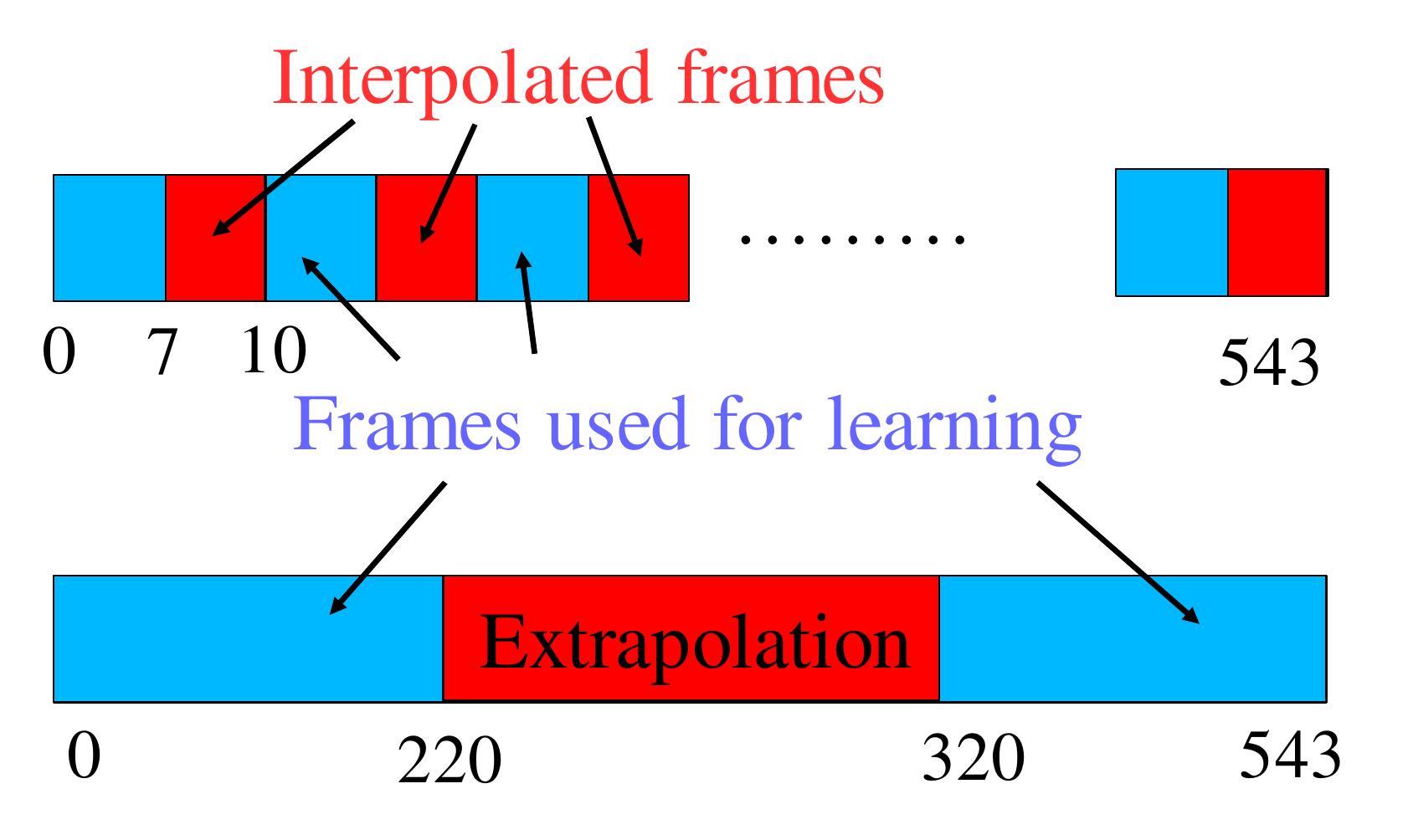}
            \caption{Frames interpolated and extrapolated for experiments.}
            \label{fig:interpolation_scenario}
            \end{minipage}

        &

        \begin{minipage}{0.33\hsize}
            \vspace{0.5cm}
                \begin{tabular}{cccc}
                \includegraphics[height=0.13\textheight]{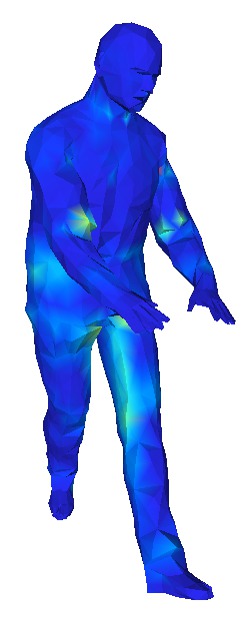} &
                \includegraphics[height=0.13\textheight]{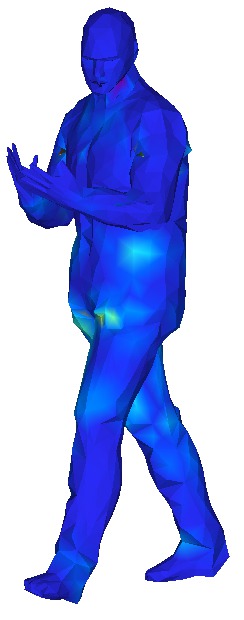} &
                \includegraphics[height=0.13\textheight]{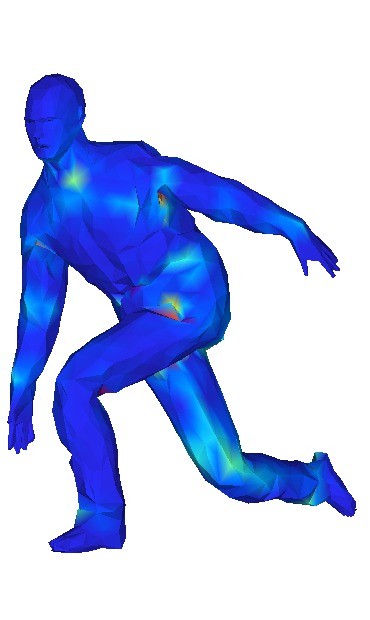} &
                \hspace{-0.3cm}

                \includegraphics[height=0.13\textheight]{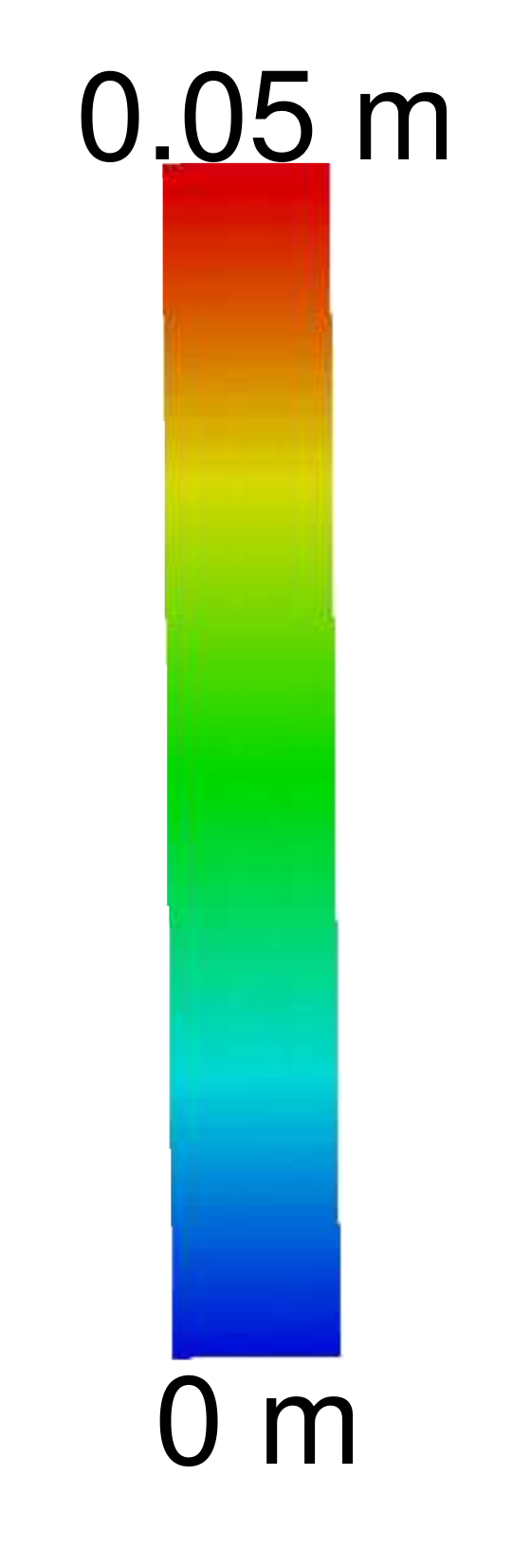}\\
            Frame \#7&
            \#247&
            \#416\\
                \end{tabular}
            \vspace{-0.3cm}
            \caption{Interpolation results by recovering 10 components by regression. RMSE 
                errors are all as small as original ones.}
            \label{fig:interpolate_body_sim}
            \vspace{-0.3cm}
            \end{minipage}

        &

        \begin{minipage}{0.33\hsize}
            \vspace{0.5cm}
                \begin{tabular}{ccc}
                \includegraphics[height=0.13\textheight]{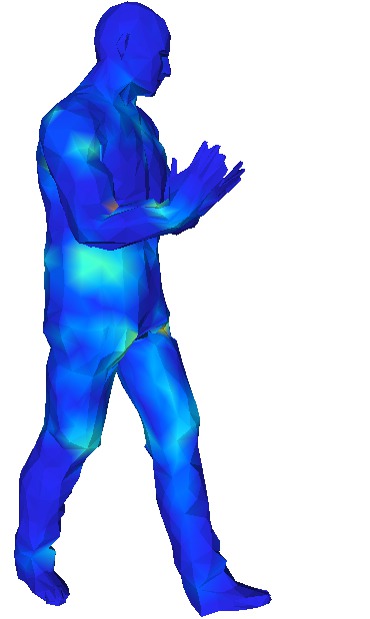} &
                \includegraphics[height=0.13\textheight]{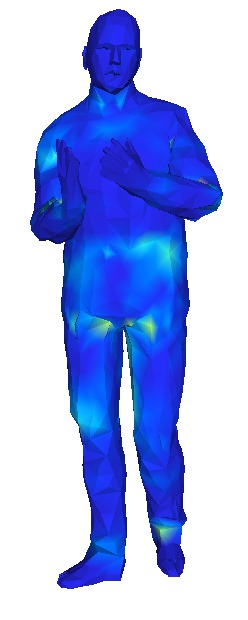} &
                \includegraphics[height=0.13\textheight]{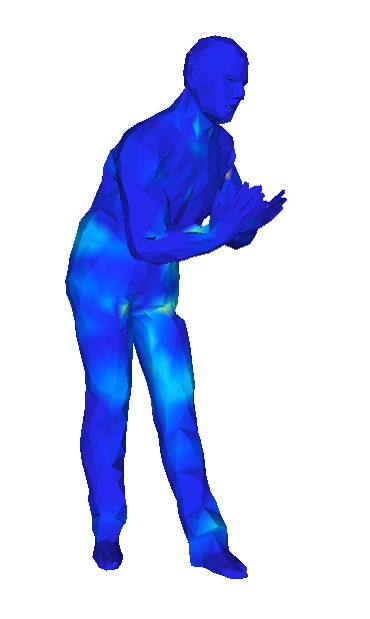} \\
            Frame \#247&
            \#280&
            \#319\\
                \end{tabular}
            \vspace{-0.3cm}
            \caption{Extrapolation results by recovering 10 components by regression. Even 
                almost new pose is reconstructed, RMSE 
                errors are still small.}
            \label{fig:extrapolate_body_sim}
            \vspace{-0.3cm}
            \end{minipage}
        \\

        \vspace{-0.3cm}
        \end{tabular}
        \end{figure*}

    \subsection{Evaluation with synthetic data}
        \vspace{-0.1cm}

        We applied our method to synthetic data for evaluation purpose.
        We use a 3D mesh model of entire human body with/without clothes, which are commercially available.
        Skeleton (bones) was also attached to the mesh model so that we were able to create sequences of 3D meshes and rendered images using 3D-CG software, \eg, 3D-MAX.
        Since muscle deformation and cloth simulation were employed in rendering, realistic shape deformations with complicated shading effects were represented.
        Some examples of rendered images are shown in \figref{SMPL}.
        In the following, the sequence of rendered images, 3D meshes with/without clothes, and poses are used for inputs.

        First, we applied eigen-texture and eigen-deformation. 
        Cumulative contribution ratios are shown in \figref{PCA_graph_sim} and differences from ground truth meshes (\ie, clothed meshes) are shown in \figref{PCA_body_sim}.
        As shown in the plots and figures, we can see that 10 eigenvectors are sufficient to represent original mesh.
        This is equivalent to 2.52\% of the original data.

        Next, we evaluated our regressors. 
        Results are shown in \figref{Regress_body_sim}.
        As shown in the figures, our NNs worked well with synthetic data and most of the body parts have small errors. 
        There are large errors around the arm joint and crotch; those errors mainly came from failure in our fitting algorithm. 
        Although those areas are usually invisible in rendered images and thus less critical for practical uses, we need to seek for a solution in our future work.

\begin{figure}[tb]
            \centering
            \includegraphics[width=\columnwidth]{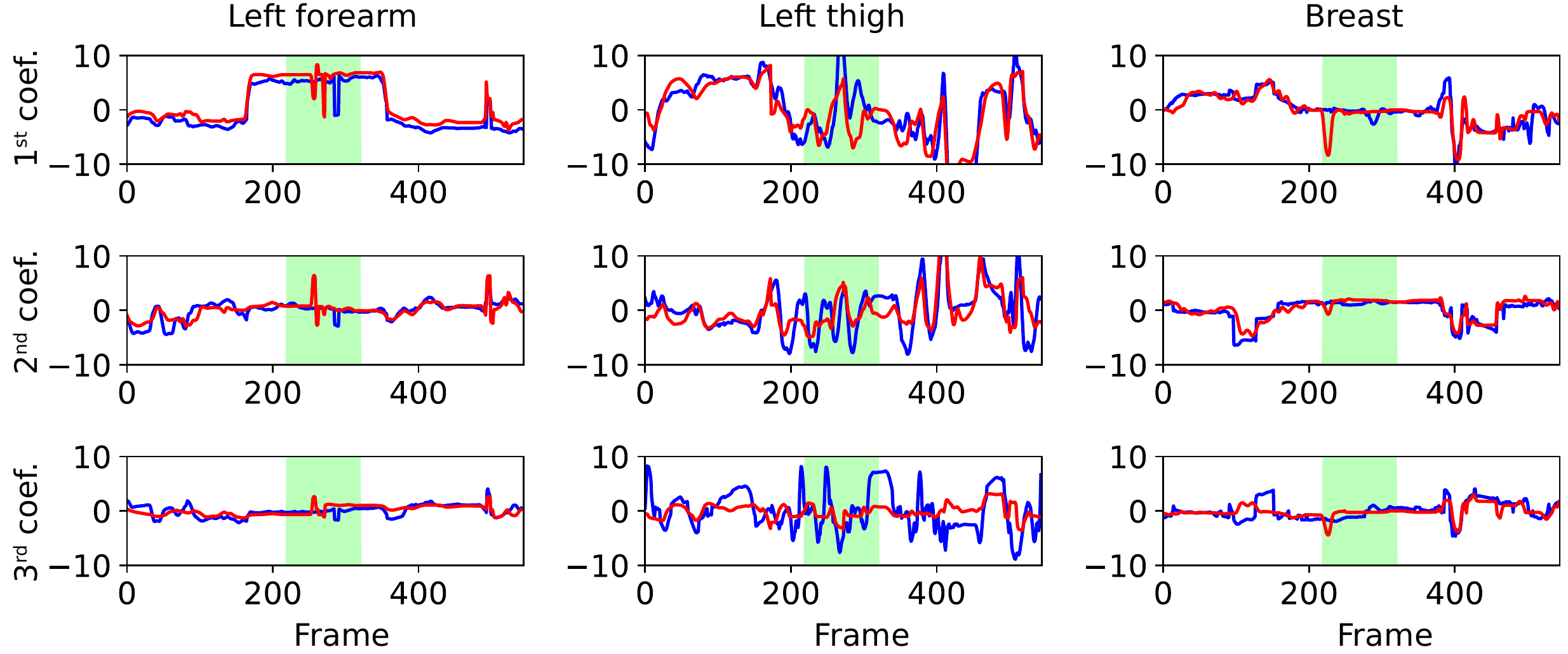}
            \vspace{-0.7cm}
            \caption{Coefficient of eigenvalue of original data and estimated data with extrapolation. The coefficients in frames 220 to 320 (highlighted) were excluded when training.}
            \label{fig:interpolate_body_sim_graph}
            \vspace{-0.3cm}
            \end{figure}

        We also evaluated interpolation accuracy. 
        We adopted two scenarios, such as short-term and long-term interpolation as explained in \figref{interpolation_scenario}. 
        Results for interpolation and extrapolation are shown in \figref{interpolate_body_sim} and \figref{extrapolate_body_sim} and coefficient values for extrapolation are shown in \figref{interpolate_body_sim_graph}.
        As shown in the figures, extrapolation tends to make larger errors than interpolation, however, regressed coefficients still have a similar trend to the ground truth.

\begin{figure}[tb]
            \begin{center}
                \tabcolsep = 0.2mm
                \begin{tabular}{cccc}
                    \rotatebox{90}{~Ground truth by CG} &
                \includegraphics[height=0.13\textheight]{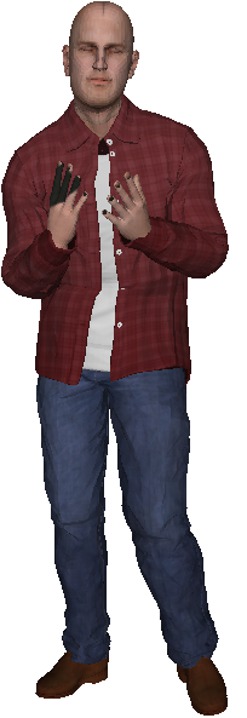} &
                \includegraphics[height=0.13\textheight]{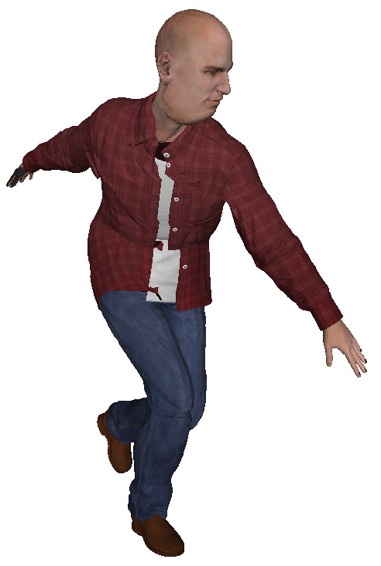} &
                \includegraphics[height=0.13\textheight]{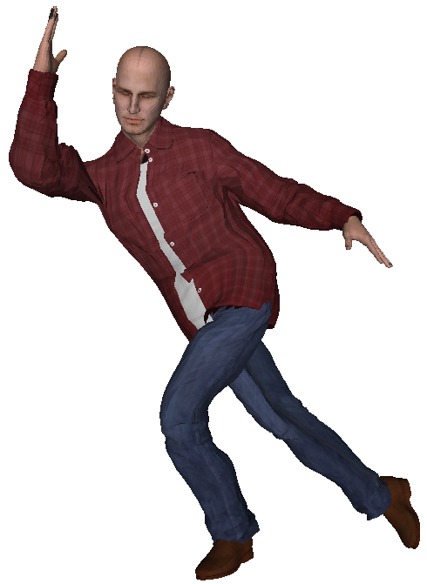} \\
                    \rotatebox{90}{~Results with texture} &
                \includegraphics[height=0.13\textheight]{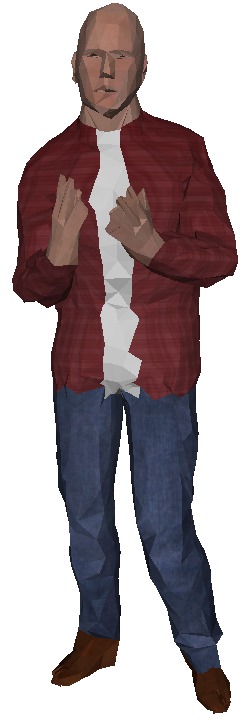} &
                \includegraphics[height=0.13\textheight]{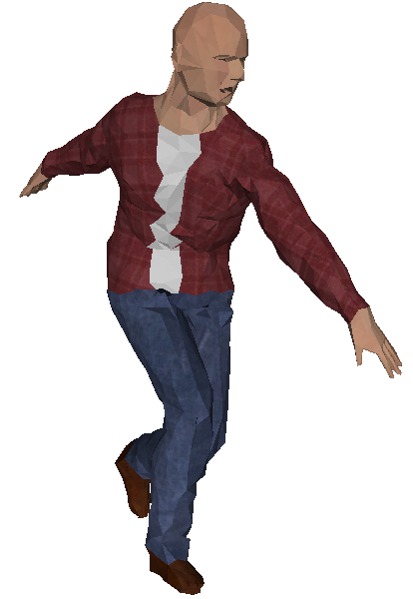} &
                \includegraphics[height=0.13\textheight]{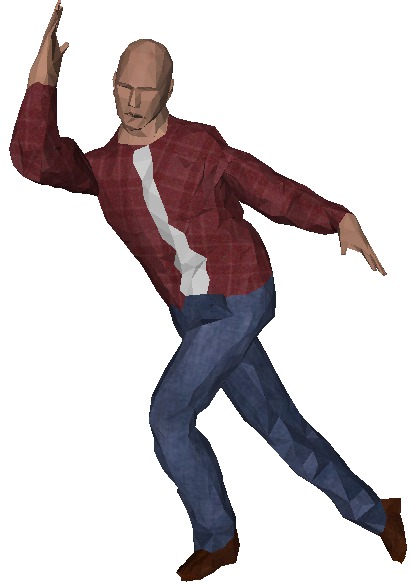} \\
            & Frame \#247 & Frame \#400 & Frame \#440 \\
                \end{tabular}
            \vspace{-.2cm}
            \caption{Input data rendered by 3D CG and final output of our method.}
            \label{fig:Texture_body_sim}
            \vspace{-.5cm}
            \end{center}
            \end{figure}

        The textured results with interpolated shapes are shown in \figref{Texture_body_sim}. 
        Considering the compression rate of 2.52\%, the visual quality of the rendered images is comparable to video compression, in addition to the significant advantage of arbitrary viewpoint rendering.

        \vspace{-0.2cm}

        \begin{figure}[t]
            \centering
            \begin{center}
                \tabcolsep = 0.8mm
                \begin{tabular}{ccccc}
                \includegraphics[height=0.13\textheight]{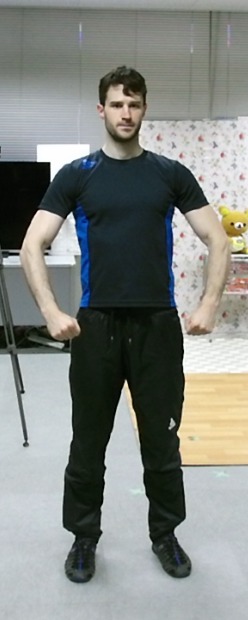} &
                \includegraphics[height=0.13\textheight]{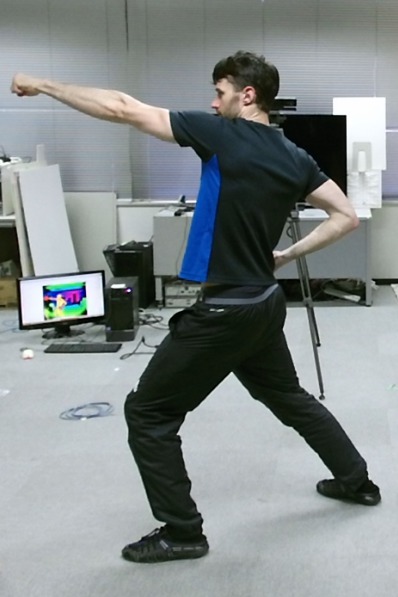} &
                \includegraphics[height=0.13\textheight]{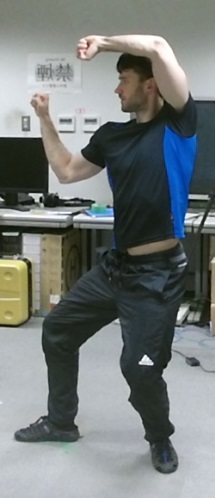} &
                \includegraphics[height=0.13\textheight]{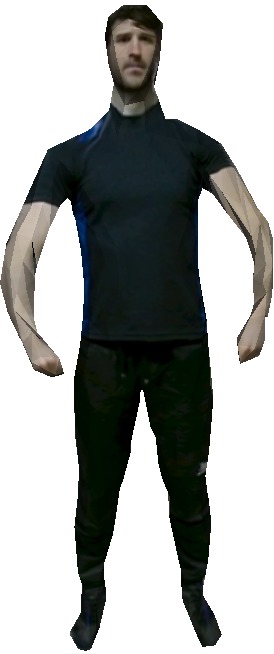} &
                \includegraphics[height=0.13\textheight]{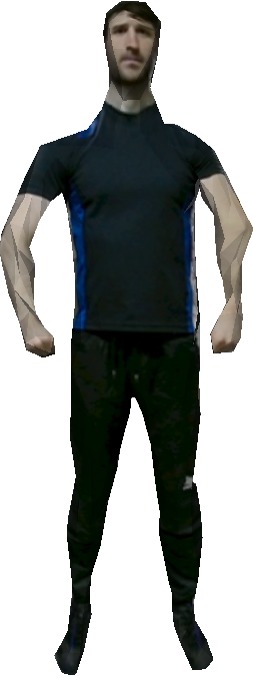} \\
                \multicolumn{3}{c}{captured images by Kinect} & Kinect & Proposed \\
            & & & \multicolumn{2}{c}{Frame \#0} \\
                \end{tabular}

                \begin{tabular}{cccc}
                \includegraphics[height=0.13\textheight]{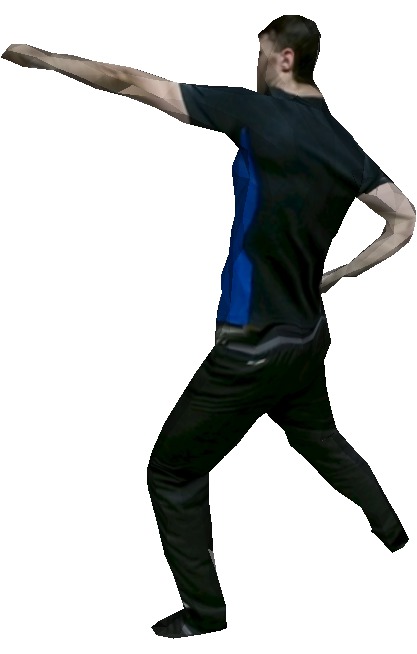} &
                \includegraphics[height=0.13\textheight]{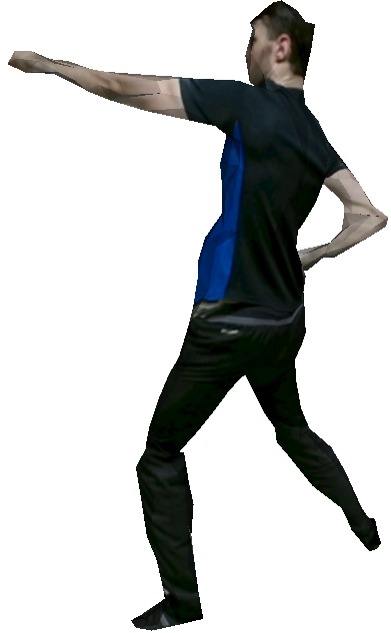} &
                \includegraphics[height=0.13\textheight]{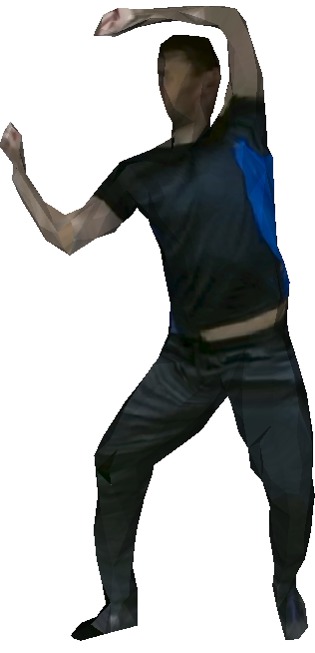} &
                \includegraphics[height=0.13\textheight]{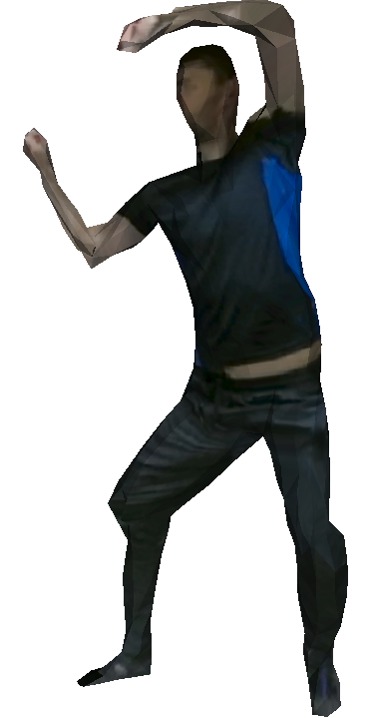} \\

            Kinect & Proposed & Kinect & Proposed \\
            \multicolumn{2}{c}{Frame \#200} &
            \multicolumn{2}{c}{Frame \#400} \\

                \end{tabular}
            \vspace{-0.2cm}
            \caption{Real input data (captured image and 3D mesh by Kinect) and the final 
                results of our method. Note that our method only uses tight clothed template 
                model, but successfully recover loose wear (pants).}
            \label{fig:Texture_body_real}
            \end{center}
            \vspace{-.8cm}
            \end{figure}

    \subsection{Demonstration with real data}
        \vspace{-0.1cm}
    
        We used two calibrated RGB-D sensors to obtain two sequences of a moving person from the person's front and back for real data experiment. 
        A pair of depth measurements from each corresponding pair of frames were integrated according to the RGB-D sensors' relative poses to make a single point cloud. 
        This point cloud, as well as the corresponding RGB images, still had unobserved surfaces due to, \eg, self-occlusion although they were not large. 
        Our system was able to synthesize the deformations in such surfaces. 
        Note that our eigen-texture and -deformation-based approach can be potentially applied even to a single RGB-D sequence. 

        Eigen-deformation results are already shown in \figref{PCA_body_real} and regression estimation results are shown in \figref{Regress_body_real}. 
        These results demonstrate that 10 eigenvectors are sufficient as well for the real data case. 
        Final rendering results are shown in \figref{Texture_body_real}, implying the effectiveness of our approach.


\vspace{-0.1cm}
\section{Conclusion}
    \vspace{-0.1cm}

    In this paper, we presented eigen-texturing and eigen-deformation method enabling full-body reconstruction with loose clothes. 
    By using lower-dimensional embeddings of texture and deformation, \ie, 10 coefficients for our datasets, the storage size required to store our model representation is drastically reduced. 
    It is also capable of long-term interpolation. 
    We evaluated our method using both synthetic and real data, proving the effectiveness of our method in both visually and quantitatively.
    In the future, more complicated shape like skirt should be taken into account.

\vspace{-0.2cm}
\section*{Acknowledgment}
\vspace{-0.2cm}
This work was supported by JSPS/KAKENHI 16H02849, 16KK0151, MIC/SCOPE 171507010 and MSR CORE12.
\vspace{-0.2cm}

\bibliographystyle{IEEEtran}
\bibliography{icpr2018.bib}

\end{document}